\documentclass[10pt,twocolumn,letterpaper]{article}

\usepackage{cvpr}
\usepackage{times}
\usepackage{epsfig}
\usepackage{graphicx}
\usepackage{amsmath}
\usepackage{amssymb}
\usepackage{color}
\usepackage{booktabs,threeparttable}
\usepackage{colortbl}
\usepackage[export]{adjustbox}
\usepackage{caption}

\usepackage[pagebackref=true,breaklinks=true,letterpaper=true,colorlinks,bookmarks=false]{hyperref}

\cvprfinalcopy % *** Uncomment this line for the final submission

\def\cvprPaperID{} % *** Enter the CVPR Paper ID here

\begin{document}

%%%%%%%%% TITLE
%\begin{CJK*}{UTF8}{song}
\title{Scaled-YOLOv4: Scaling Cross Stage Partial Network}

\author{Chien-Yao Wang\\
Institute of Information Science\\
Academia Sinica, Taiwan\\
{\tt\small kinyiu@iis.sinica.edu.tw}
\and
Alexey Bochkovskiy\\
{\tt\small alexeyab84@gmail.com}
\and
Hong-Yuan Mark Liao\\
Institute of Information Science\\
Academia Sinica, Taiwan\\
{\tt\small liao@iis.sinica.edu.tw}
}

\maketitle
%\thispagestyle{empty}

%%%%%%%%% ABSTRACT
\begin{abstract}
	We show that the YOLOv4 object detection neural network based on the CSP approach, scales both up and down and is applicable to small and large networks while maintaining optimal speed and accuracy. We propose a network scaling approach that modifies not only the depth, width, resolution, but also structure of the network. YOLOv4-large model achieves state-of-the-art results: 55.5\% AP (73.4\% AP$_{50}$) for the MS COCO dataset at a speed of  $\sim$16 FPS on Tesla V100, while with the test time augmentation, YOLOv4-large achieves 56.0\% AP (73.3 AP$_{50}$). To the best of our knowledge, this is currently the highest accuracy on the COCO dataset among any published work. The YOLOv4-tiny model achieves 22.0\% AP (42.0\% AP$_{50}$) at a speed of $\sim$443 FPS on RTX 2080Ti, while by using TensorRT, batch size = 4 and FP16-precision the YOLOv4-tiny achieves 1774 FPS.
\end{abstract}

%%%%%%%%% BODY TEXT
%-------------------------------------------------------------------------
\section{Introduction}

The deep learning-based object detection technique has many applications in our daily life. For example, medical image analysis, self-driving vehicles, business analytics, and face identification all rely on object detection.  The computing facilities required for the above applications maybe cloud computing facilities, general GPU, IoT clusters, or single embedded device.  In order to design an effective object detector, model scaling technique is very important, because it can make object detector achieve high accuracy and real-time inference on various types of devices.

\begin{figure}[t]
	\begin{center}
		\includegraphics[width=1.0\linewidth]{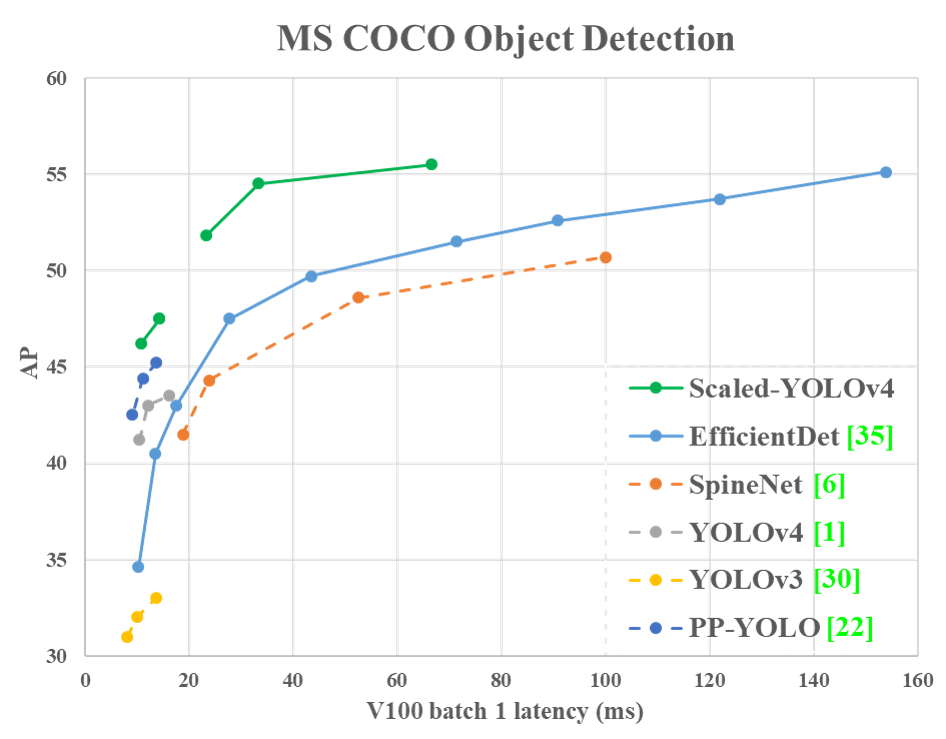}
	\end{center}
	\caption{Comparison of the proposed YOLOv4 and other state-of-the-art object detectors. The dashed line means only latency of model inference, while the solid line include model inference and post-processing.}
	\label{fig:ap}
\end{figure}

The most common model scaling technique is to change the depth (number of convolutional layers in a CNN) and width (number of convolutional filters in a convolutional layer) of the backbone, and then train CNNs suitable for different devices.  For example among the ResNet \cite{he2016deep} series, ResNet-152 and ResNet-101 are often used in cloud server GPUs, ResNet-50 and ResNet-34 are often used in personal computer GPUs, and ResNet-18 and ResNet-10 can be used in low-end embedded systems.  In \cite{cai2019once}, Cai \textit{et al.} try to develop techniques that can be applied to various device network architectures with only training once.  They use techniques such as decoupling training and search and knowledge distillation to decouple and train several sub-nets, so that the entire network and sub-nets are capable of processing target tasks.  Tan \textit{et al.} \cite{tan2019efficientnet} proposed using NAS technique to perform compound scaling, including the treatment of width, depth, and resolution on EfficientNet-B0. They use this initial network to search for the best CNN architecture for a given amount of computation and set it as EfficientNet-B1, and then use linear scale-up technique to obtain architectures such as EfficientNet-B2 to EfficientNet-B7.  Radosavovic \textit{et al.} \cite{radosavovic2020designing} summarized and added constraints from the vast parameter search space AnyNet, and then designed RegNet.  In RegNet, they found that the optimal depth of CNN is about 60.  They also found when the bottleneck ratio is set to 1 and the width increase rate of cross-stage is set to 2.5 will receive the best performance.  In addition, recently there are NAS and model scaling methods specifically proposed for object detection, such as SpineNet \cite{du2019spinenet} and EfficientDet \cite{tan2019efficientdet}.

Through analysis of state-of-the-art object detectors \cite{bochkovskiy2020yolov4, cao2020d2det, du2019spinenet, qiu2020borderdet, tan2019efficientdet, wang2020scale, zhang2020dynamic}, we found that CSPDarknet53, which is the backbone of YOLOv4 \cite{bochkovskiy2020yolov4}, matches almost all optimal architecture features obtained by network architecture search technique.  The depth of CSPDarknet53, bottleneck ratio, width growth ratio between stages are 65, 1, and 2, respectively.  Therefore, we developed model scaling technique based on YOLOv4 and proposed scaled-YOLOv4.  The proposed scaled-YOLOv4 turned out with excellent performance, as illustrated in Figure \ref{fig:ap}.  The design procedure of scaled-YOLOv4 is as follows.  First, we re-design YOLOv4 and propose YOLOv4-CSP, and then based on YOLOv4-CSP we developed scaled-YOLOv4.  In the proposed scaled-YOLOv4, we discussed the upper and lower bounds of linear scaling up/down models, and respectively analyzed the issues that need to be paid attention to in model scaling for small models and large models.  Thus, we are able to systematically develop YOLOv4-large and YOLOv4-tiny models.  Scaled-YOLOv4 can achieve the best trade-off between speed and accuracy, and is able to perform real-time object detection on 16 FPS, 30 FPS, and 60 FPS movies, as well as embedded systems.

We summarize the contributions of this paper : (1) design a powerful model scaling method for small model, which can systematically balance the computation cost and memory bandwidth of a shallow CNN; (2) design a simple yet effective strategy for scaling a large object detector; (3) analyze the relations among all model scaling factors and then perform model scaling based on most advantageous group partitions; (4) experiments have confirmed that the FPN structure is inherently a once-for-all structure; and (5) we make use of the above methods to develop YOLOv4-tiny and YOLO4v4-large.

%-------------------------------------------------------------------------
\section{Related work}

\subsection{Real-time object detection}

Object detectors is mainly divided into one-stage object detectors \cite{redmon2016you, redmon2017yolo9000, redmon2018yolov3, liu2016ssd, lin2017focal, qiao2020detectors} and two-stage object detectors \cite{girshick2014rich, girshick2015fast, ren2015faster}.  The output of one-stage object detector can be obtained after only one CNN operation.  As for two-stage object detector, it usually feeds the high score region proposals obtained from the first-stage CNN to the second-stage CNN for final prediction.  The inference time of one-stage object detectors and two-stage object detectors can be expressed as ${T}_{one} = {T}_{{1}^{st}}$ and ${T}_{two} = {T}_{{1}^{st}} + m{T}_{{2}^{nd}}$, where $m$ is the number of region proposals whose confidence score is higher than a threshold.  In other words, the inference time required for one-stage object detectors is constant, while the inference time required for two-stage object detectors is not fixed.  So if we need real-time object detectors, they are almost necessarily one-stage object detectors.  Today’s popular one-stage object detectors mainly have two kinds: anchor-based \cite{redmon2018yolov3, lin2017focal} and anchor-free \cite{duan2019centernet, law2018cornernet, law2019cornernet, tian2019fcos}.  Among all anchor-free approaches, CenterNet \cite{zhou2019objects} is very popular because it does not require complicated post-processing, such as Non-Maximum Suppression (NMS).  At present, the more accurate real-time one-stage object detectors are anchor-based EfficientDet \cite{tan2019efficientdet}, YOLOv4 \cite{bochkovskiy2020yolov4}, and PP-YOLO \cite{long2020pp}.  In this paper, we developed our model scaling methods based on YOLOv4 \cite{bochkovskiy2020yolov4}.

\subsection{Model scaling}

Traditional model scaling method is to change the depth of a model, that is to add more convolutional layers. For example, the VGGNet \cite{simonyan2014very} designed by Simonyan \textit{et al.} stacks additional convolutional layers in different stages, and also uses this concept to design VGG-11, VGG-13, VGG-16, and VGG-19 architectures.  The subsequent methods generally follow the same methodology for model scaling.  For the ResNet \cite{he2016deep} proposed by He \textit{et al.}, depth scaling can construct very deep networks, such as ResNet-50, ResNet-101, and ResNet-152.  Later, Zagoruyko \textit{et al.} \cite{zagoruyko2016wide} thought about the width of the network, and they changed the number of kernel of convolutional layer to realize scaling. They therefore design wide ResNet (WRN) , while maintaining the same accuracy. Although WRN has higher amount of parameters than ResNet, the inference speed is much faster. The subsequent DenseNet \cite{huang2017densely} and ResNeXt \cite{xie2017aggregated} also designed a compound scaling version that puts depth and width into consideration. As for image pyramid inference, it is a common way to perform augmentation at run time.  It takes an input image and makes a variety of different resolution scaling, and then input these distinct pyramid combinations into a trained CNN.  Finally, the network will integrate the multiple sets of outputs as its ultimate outcome.  Redmon \textit{et al.} \cite{redmon2018yolov3} use the above concept to execute input image size scaling.  They use higher input image resolution to perform fine-tune on a trained Darknet53, and the purpose of executing this step is to get higher accuracy.   

In recent years, network architecture search (NAS) related research has been developed vigorously, and NAS-FPN \cite{ghiasi2019fpn} has searched for the combination path of feature pyramid.  We can think of NAS-FPN as a model scaling technique which is mainly executed at the stage level.  As for EfficientNet \cite{tan2019efficientnet}, it uses compound scaling search based on depth, width, and input size.  The main design concept of EfficientDet \cite{tan2019efficientdet} is to disassemble the modules with different functions of object detector, and then perform scaling on the image size, width, \#BiFPN layers, and \#box/class layer.  Another design that uses NAS concept is SpineNet \cite{du2019spinenet}, which is mainly aimed at the overall architecture of fish-shaped object detector for network architecture search.  This design concept can ultimately produce a scale-permuted structure.  Another network with NAS design is RegNet \cite{radosavovic2020designing}, which mainly fixes the number of stage and input resolution, and integrates all parameters such as depth, width, bottleneck ratio and group width of each stage into depth, initial width, slope, quantize, bottleneck ratio, and group width.  Finally, they use these six parameters to perform compound model scaling search.  The above methods are all great work, but few of them analyze the relation between different parameters.  In this paper, we will try to find a method for synergistic compound scaling based on the design requirements of object detection.

%------------------------------------------------------------------------
\section{Principles of model scaling}

After performing model scaling for the proposed object detector, the next step is to deal with the quantitative factors that will change, including the number of parameters with qualitative factors.  These factors include model inference time, average precision, etc.  The qualitative factors will have different gain effects depending on the equipment or database used.  We will analyze and design for quantitative factors in \ref{ss:gp}.  As for \ref{ss:tp} and \ref{ss:lp}, we will design qualitative factors related to tiny object detector running on low-end device and high-end GPUs respectively.

\subsection{General principle of model scaling}
\label{ss:gp}

When designing the efficient model scaling methods, our main principle is that when the scale is up/down, the lower/higher the quantitative cost we want to increase/decrease, the better.  In this section, we will show and analyze various general CNN models, and try to understand their quantitative costs when facing changes in (1) image size, (2) number of layers, and (3) number of channels.  The CNNs we chose are ResNet, ResNext, and Darknet.

For the $k$-layer CNNs with $b$ base layer channels, the computations of ResNet layer is $k*$[conv($1\times1$, $b/4$) $\rightarrow$ conv($3\times3$, $b/4$) $\rightarrow$ conv($1\times1$, $b$)], and that of ResNext layer is $k*$[conv($1\times1$, $b/2$) $\rightarrow$ gconv($3\times3/32$, $b/2$) $\rightarrow$ conv($1\times1$, $b$)].  As for the Darknet layer, the amount of computation is $k*$[conv($1\times1$, $b/2$) $\rightarrow$ conv($3\times3$, $b$)].  Let the scaling factors that can be used to adjust the image size, the number of layers, and the number of channels be $\alpha$, $\beta$, and $\gamma$, respectively.  When these scaling factors vary, the corresponding changes on FLOPs are summarized in Table \ref{table:t1}.

\begin{table}[h]
	\centering
	\begin{threeparttable}[h]
		\footnotesize
		\caption{FLOPs of different computational layers with different model scalng factors.}
		\label{table:t1}
		\setlength\tabcolsep{3.5pt}
		\begin{tabular}{llccc}
			\toprule
			\textbf{Model} & \textbf{original} & \textbf{size $\alpha$} & \textbf{depth $\beta$} & \textbf{width $\gamma$} \\			
			\midrule
			Res layer & $r = 17whk{b}^{2}/16$ & ${\alpha}^{2}r$ & ${\beta}r$ & ${\gamma}^{2}r$ \\
			ResX layer & $x = 137whk{b}^{2}/128$ & ${\alpha}^{2}x$ & ${\beta}x$ & ${\gamma}^{2}x$ \\
			Dark layer & $d = 5whk{b}^{2}$ & ${\alpha}^{2}d$ & ${\beta}d$ & ${\gamma}^{2}d$ \\
			\bottomrule
		\end{tabular}
	\end{threeparttable}
\end{table}

It can be seen from Table \ref{table:t1} that the scaling size, depth, and width cause increase in the computation cost.  They respectively show square, linear, and square increase.

The CSPNet \cite{wang2020cspnet} proposed by Wang \textit{et al.} can be applied to various CNN architectures, while reducing the amount of parameters and computations. In addition, it also improves accuracy and reduces inference time.  We apply it to ResNet, ResNeXt, and Darknet and observe the changes in the amount of computations, as shown in Table \ref{table:t3}.

\begin{table}[h]
	\centering
	\begin{threeparttable}[h]
		\footnotesize
		\caption{FLOPs of different computational layers with/without CSP-ization.}
		\label{table:t3}
		\setlength\tabcolsep{3.5pt}
		\begin{tabular}{lcc}
			\toprule
			\textbf{Model} & \textbf{original} & \textbf{to CSP} \\				
			\midrule
			Res layer & $17whk{b}^{2}/16$ & $wh{b}^{2}(3/4+13k/16)$ \\
			ResX layer & $137whk{b}^{2}/128$ & $wh{b}^{2}(3/4+73k/128)$ \\
			Dark layer & $5whk{b}^{2}$ & $wh{b}^{2}(3/4+5k/2)$ \\
			\bottomrule
		\end{tabular}
	\end{threeparttable}
\end{table}

From the figures shown in Table \ref{table:t3}, we observe that after converting the above CNNs to CSPNet, the new architecture can effectively reduce the amount of computations (FLOPs) on ResNet, ResNeXt, and Darknet by 23.5\%, 46.7\%, and 50.0\%, respectively.  Therefore, we use CSP-ized models as the best model for performing model scaling.

\subsection{Scaling Tiny Models for Low-End Devices}
\label{ss:tp}

For low-end devices, the inference speed of a designed model is not only affected by the amount of computation and model size, but more importantly, the limitation of peripheral hardware resources must be considered.  Therefore, when performing tiny model scaling, we must also consider factors such as memory bandwidth, memory access cost (MACs), and DRAM traffic.  In order to take into account the above factors, our design must comply with the following principles:

\noindent
\textbf{Make the order of computations less than $O(whk{b}^{2})$:} Lightweight models are different from large models in that their parameter utilization efficiency must be higher in order to achieve the required accuracy with a small amount of computations.  When performing model scaling, we hope the order of computation can be as low as possible.  In Table \ref{table:t4}, we analyze the network with efficient parameter utilization, such as the computation load of DenseNet and OSANet \cite{lee2019energy}, where $g$ means growth rate.

\begin{table}[h]
	\centering
	\begin{threeparttable}[h]
		\footnotesize
		\caption{FLOPs of Dense layer and OSA layer.}
		\label{table:t4}
		\setlength\tabcolsep{3.5pt}
		\begin{tabular}{lc}
			\toprule
			\textbf{Model} & \textbf{FLOPs} \\				
			\midrule
			Dense layer & $whgbk + wh{g}^{2}k(k-1)/2$ \\
			OSA layer & $whbg + wh{g}^{2}(k-1)$ \\
			\bottomrule
		\end{tabular}
	\end{threeparttable}
\end{table}

For general CNNs, the relationship among $g$, $b$, and $k$ listed in Table \ref{table:t4} is $k << g < b$.  Therefore, the order of computation complexity of DenseNet is $O(whgbk)$, and that of OSANet is $O(max(whbg,whk{g}^{2}))$.  The order of computation complexity of the above two is less than $O(whk{b}^{2})$ of the ResNet series.  Therefore, we design our tiny model with the help of OSANet, which has a smaller computation complexity.

\noindent
\textbf{Minimize/balance size of feature map:} 
In order to get the best trade-off in terms of computing speed, we propose a new concept, which is to perform gradient truncation between computational block of the CSPOSANet.  If we apply the original CSPNet design to the DenseNet or ResNet architectures, because the ${j}^{th}$ layer output of these two architectures is the integration of the ${1}^{st}$ to ${(j-1)}^{th}$ layer outputs, we must treat the entire computational block as a whole.  Because the computational block of OSANet belongs to the PlainNet architecture, making CSPNet from any layer of a computational block can achieve the effect of gradient truncation.  We use this feature to re-plan the $b$ channels of the base layer and the $kg$ channels generated by computational block, and split them into two paths with equal channel numbers, as shown in Table \ref{table:t5}.

\begin{table}[h]
	\centering
	\begin{threeparttable}[h]
		\footnotesize
		\caption{Number of channel of OSANet, CSPOSANet, and CSPOSANet with partial in computational block (PCB).}
		\label{table:t5}
		\setlength\tabcolsep{7.5pt}
		\begin{tabular}{llll}
			\toprule
			\textbf{layer ID} & \textbf{original} & \textbf{CSP} & \textbf{partial in CB} \\				
			\midrule
			1 & $b \rightarrow g$ & $g \rightarrow g$ & $g \rightarrow g$ \\
			2 & $g \rightarrow g$ & $g \rightarrow g$ & $g \rightarrow g$ \\
			... & $g \rightarrow g$ & $g \rightarrow g$ & $g \rightarrow g$ \\
			$k$ & $g \rightarrow g$ & $g \rightarrow g$ & $g \rightarrow g$ \\
			$T$ & \begin{tabular}{@{}l@{}}$(b+kg)$ \\ $\rightarrow (b+kg)/2$\end{tabular} & $kg \rightarrow kg$ &   \begin{tabular}{@{}l@{}}$(b+kg)/2$ \\ $\rightarrow (b+kg)/2$\end{tabular}\\
			\bottomrule
		\end{tabular}
	\end{threeparttable}
\end{table}

When the number of channel is $b+kg$, if one wants to split these channels into two paths, the best partition is to divide it into two equal parts, i.e. $(b+kg)/2$.  When we actually consider the bandwidth $\tau$ of the hardware, if software optimization is not considered, the best value is $ceil((b+kg)/2\tau)\times\tau$.  The CSPOSANet we designed can dynamically adjust the channel allocation.

\noindent
\textbf{Maintain the same number of channels after convolution:} For evaluating the computation cost of low-end device, we must also consider power consumption, and the biggest factor affecting power consumption is memory access cost (MAC).  Usually the MAC calculation method for a convolution operation is as follows:

\begin{equation}
MAC = hw({C}_{in}+{C}_{out})+K{C}_{in}{C}_{out}
\end{equation}
where $h$, $w$, ${C}_{in}$, ${C}_{out}$, and $K$ represent, respectively, the height and width of feature map, the channel number of input and output, and the kernel size of convolutional filter.  By calculating geometric inequalities, we can derive the smallest MAC when ${C}_{in}={C}_{out}$ \cite{ma2018shufflenetv2}.

\noindent
\textbf{Minimize Convolutional Input/Output (CIO):} CIO \cite{chao2019hardnet} is an indicator that can measure the status of DRAM IO.  Table \ref{table:t6} lists the CIO of OSA, CSP, and our designed CSPOSANet.

\begin{table}[h]
	\centering
	\begin{threeparttable}[h]
		\footnotesize
		\caption{CIO of OSANet, CSPOSANet, and CSPOSANet with PCB.}
		\label{table:t6}
		\setlength\tabcolsep{3.5pt}
		\begin{tabular}{lll}
			\toprule
			\textbf{original} & \textbf{CSP} & \textbf{partial in CB} \\				
			\midrule
			$bg + (k-1){g}^{2} + {(b+kg)}^{2}/2$ & $k{g}^{2} + {(kg)}^{2}$ & $k{g}^{2} + {(b+kg)}^{2}/4$ \\
			\bottomrule
		\end{tabular}
	\end{threeparttable}
\end{table}

When $kg > b/2$, the proposed CSPOSANet can obtain the best CIO.

\subsection{Scaling Large Models for High-End GPUs}
\label{ss:lp}

Since we hope to improve the accuracy and maintain the real-time inference speed after scaling up the CNN model, we must find the best combination among the many scaling factors of object detector when performing compound scaling.  Usually, we can adjust the scaling factors of an object detector’s input, backbone, and neck.  The potential scaling factors that can be adjusted are summarized as Table \ref{table:t7}.

\begin{table}[h]
	\centering
	\begin{threeparttable}[h]
		\footnotesize
		\caption{Model scaling factors of different parts of object detectors.}
		\label{table:t7}
		\setlength\tabcolsep{3.5pt}
		\begin{tabular}{ll}
			\toprule
			\textbf{Part} & \textbf{Scaling Factor} \\				
			\midrule
			Input & size$^{input}$ \\
			Backbone & width$^{backbone}$ , depth$^{backbone}$, \#stage$^{backbone}$ \\
			Neck & width$^{neck}$, depth$^{neck}$, \#stage$^{neck}$  \\
			\bottomrule
		\end{tabular}
	\end{threeparttable}
\end{table}

The biggest difference between image classification and object detection is that the former only needs to identify the category of the largest component in an image, while the latter needs to predict the position and size of each object in an image.  In one-stage object detector, the feature vector corresponding to each location is used to predict the category and size of an object at that location.  The ability to better predict the size of an object basically depends on the receptive field of the feature vector.  In the CNN architecture, the thing that is most directly related to receptive field is the stage, and the feature pyramid network (FPN) architecture tells us that higher stages are more suitable for predicting large objects.  In Table \ref{table:t8}, we illustrate the relations between receptive field and several parameters.

\begin{table}[h]
	\centering
	\begin{threeparttable}[h]
		\footnotesize
		\caption{Effect of receptive field caused by different model scaling factors.}
		\label{table:t8}
		\setlength\tabcolsep{3.5pt}
		\begin{tabular}{ll}
			\toprule
			\textbf{Scaling factor} & \textbf{Effect of receptive field} \\				
			\midrule
			size$^{input}$ & no effect. \\
			width & no effect. \\
			depth & one more $k \times k$ conv layer, increases $k-1$. \\
			\#stage & one more stage, receptive field doubled. \\
			\bottomrule
		\end{tabular}
	\end{threeparttable}
\end{table}

From Table \ref{table:t8}, it is apparent that width scaling can be independently operated.  When the input image size is increased, if one wants to have a better prediction effect for large objects, he/she must increase the depth or number of stages of the network.  Among the parameters listed in Table \ref{table:t8}, the compound of \{size$^{input}$, \#stage\} turns out with the best impact.  Therefore, when performing scaling up, we first perform compound scaling on {size$^{input}$, \#stage}, and then according to real-time requirements, we further perform scaling on depth and width respectively.

%------------------------------------------------------------------------
\section{Scaled-YOLOv4}

In this section, we put our emphasis on designing scaled YOLOv4 for general GPUs, low-end GPUs, and high-end GPUs.

\subsection{CSP-ized YOLOv4}

YOLOv4 is designed for real-time object detection on general GPU.  In this sub-section, we re-design YOLOv4 to YOLOv4-CSP to get the best speed/accuracy trade-off.

\noindent
\textbf{Backbone:} In the design of CSPDarknet53, the computation of down-sampling convolution for cross-stage process is not included in a residual block.  Therefore, we can deduce that the amount of computation of each CSPDarknet stage is $wh{b}^{2}(9/4+3/4+5k/2)$.  From the formula deduced above, we know that CSPDarknet stage will have a better computational advantage over Darknet stage only when $k > 1$ is satisfied.  The number of residual layer owned by each stage in CSPDarknet53 is 1-2-8-8-4 respectively.  In order to get a better speed/accuracy trade-off, we convert the first CSP stage into original Darknet residual layer.

\begin{figure}[h]
	\begin{center}
		\includegraphics[width=1.0\linewidth]{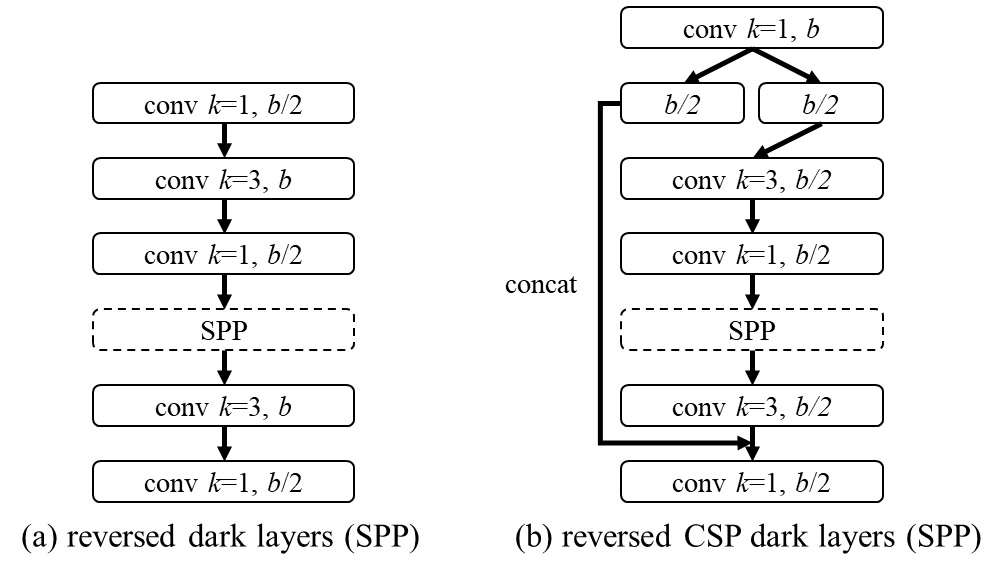}
	\end{center}
	\vspace{-4mm}
	\caption{Computaional blocks of reversed Dark layer (SPP) and reversed CSP dark layers (SPP).}
	\label{fig:rcsp}
\end{figure}

\noindent
\textbf{Neck:} In order to effectively reduce the amount of computation, we CSP-ize the PAN \cite{liu2018path} architecture in YOLOv4.  The computation list of a PAN architecture is illustrated in Figure \ref{fig:rcsp}(a).  It mainly integrates the features coming from different feature pyramids, and then passes through two sets of reversed Darknet residual layer without shortcut connections.  After CSP-ization, the architecture of the new computation list is shown in Figure \ref{fig:rcsp}(b). This new update effectively cuts down 40\% of computation.

\noindent
\textbf{SPP:} The SPP module was originally inserted in the middle position of the first computation list group of the neck.  Therefore, we also inserted SPP module in the middle position of the first computation list group of the CSPPAN.

\subsection{YOLOv4-tiny}

YOLOv4-tiny is designed for low-end GPU device, the design will follow principles mentioned in section \ref{ss:tp}.

\begin{figure}[h]
	\begin{center}
		\includegraphics[width=0.7\linewidth]{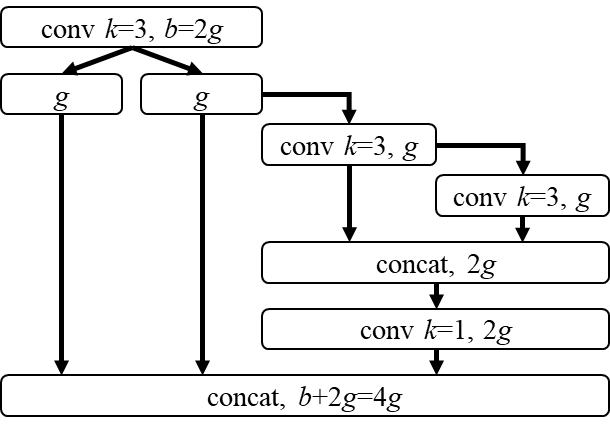}
	\end{center}
	\vspace{-4mm}
	\caption{Computational block of YOLOv4-tiny.}
	\label{fig:pcb}
\end{figure}

We will use the CSPOSANet with PCB architecture to form the backbone of YOLOv4. We set $g = b/2$ as the growth rate and make it grow to $b/2+kg=2b$ at the end.  Through calculation, we deduced $k = 3$, and its architecture is shown in Figure \ref{fig:pcb}.  As for the number of channels of each stage and the part of neck, we follow the design of YOLOv3-tiny.

\begin{figure*}[t]
	\begin{center}
		\includegraphics[width=1.0\linewidth]{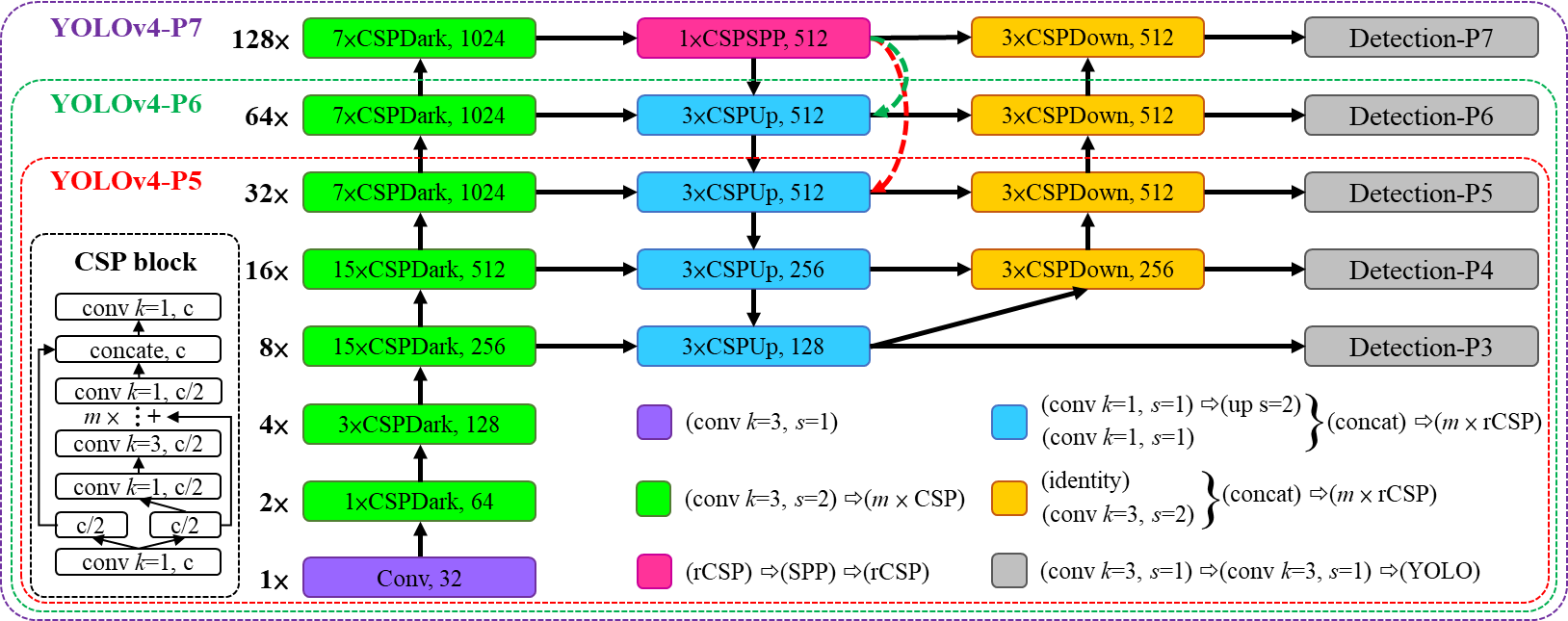}
	\end{center}
	\vspace{-4mm}
	\caption{Architecture of YOLOv4-large, including YOLOv4-P5, YOLOv4-P6, and YOLOv4-P7. The dashed arrow means replace the corresponding CSPUp block by CSPSPP block.}
	\label{fig:yolov4-large}
\end{figure*}

\subsection{YOLOv4-large}

YOLOv4-large is designed for cloud GPU, the main purpose is to achieve high accuracy for object detection. We designed a fully CSP-ized model YOLOv4-P5 and scaling it up to YOLOv4-P6 and YOLOv4-P7.

Figure \ref{fig:yolov4-large} shows the structure of YOLOv4-P5, YOLOv4-P6, and YOLOv4-P7.  We designed to perform compound scaling on {size$^{input}$, \#stage}.  We set the depth scale of each stage to ${2}^{{d}_{{s}_{i}}}$, and ${d}_{s}$ to [1, 3, 15, 15, 7, 7, 7].  Finally, we further use inference time as constraint to perform additional width scaling.  Our experiments show that YOLOv4-P6 can reach real-time performance at 30 FPS video when the width scaling factor is equal to 1.  For YOLOv4-P7, it can reach real-time performance at 16 FPS video when the width scaling factor is equal to 1.25.

%-------------------------------------------------------------------------
\section{Experiments}

We use MSCOCO 2017 object detection dataset to verify the proposed scaled-YOLOv4.  We do not use ImageNet pre-trained models, and all scaled-YOLOv4 models are trained from scratch and the adopted tool is SGD optimizer.  The time used for training YOLOv4-tiny is 600 epochs, and that used for training YOLOv4-CSP is 300 epochs.  As for YOLOv4-large, we execute 300 epochs first and then followed by using stronger data augmentation method to train 150 epochs.  As for the Lagrangian multiplier of hyper-parameters, such as anchors of learning rate, the degree of different data augmentation methods, we use k-means and genetic algorithms to determine.  All details related to hyper-parameters are elaborated in Appendix.

\subsection{Ablation study on CSP-ized model}

In this sub-section, we will CSP-ize different models and analyze the impact of CSP-ization on the amount of parameters, computations, throughput, and average precision.  We use Darknet53 (D53) as backbone and choose FPN with SPP (FPNSPP) and PAN with SPP (PANSPP) as necks to design ablation studies.  In Table \ref{table:t9} we list the AP$^{val}$ results after CSP-izing different DNN models.  We use LeakyReLU (Leaky) and Mish activation function respectively to compare the amount of used parameters, computations, and throughput.  Experiments are all conducted on COCO minval dataset and the resulting APs are shown in the last column of Table \ref{table:t9}.

\begin{table}[h]
	\centering
	\begin{threeparttable}[h]
		\footnotesize
		\caption{Ablation study of CSP-ized models @608$\times$608.}
		\label{table:t9}
		\setlength\tabcolsep{3.5pt}
		\begin{tabular}{llccccc}
			\toprule
			\textbf{Backbone} & \textbf{Neck} & \textbf{Act.} & \textbf{\#Param.} & \textbf{FLOPs} & \begin{tabular}{@{}c@{}}\textbf{Batch} \\ \textbf{8 FPS}\end{tabular} & \textbf{AP$^{val}$} \\				
			\midrule
			D53 & FPNSPP & Leaky & 63M & 142B & 208 & 43.5\% \\
			D53 & FPNSPP & Mish & 63M & 142B & 196 & 45.3\% \\
			\textbf{CD53s} & \textbf{CFPNSPP} & Leaky & 43M & 97B & 222 & 45.7\% \\
			\textbf{CD53s} & \textbf{CFPNSPP} & Mish & 43M & 97B & 208 & 46.3\% \\
			D53 & PANSPP & Leaky & 78M & 160B & 196 & 46.5\% \\
			D53 & PANSPP & Mish & 78M & 160B & 185 & 46.9\% \\
			\textbf{CD53s} & \textbf{CPANSPP} & Leaky & 53M & 109B & 208 & 46.9\% \\
			\textbf{CD53s} & \textbf{CPANSPP} & Mish & 53M & 109B & 200 & 47.5\% \\
			\bottomrule
		\end{tabular}
	\end{threeparttable}
\end{table}

From the data listed in Table \ref{table:t9}, it can be seen that the CSP-ized models have greatly reduced the amount of parameters and computations by 32\%, and brought improvements in both Batch 8 throughput and AP.  If one wants to maintain the same frame rate, he/she can add more layers or more advanced activation functions to the models after CSP-ization.  From the figures shown in Table \ref{table:t9}, we can see that both CD53s-CFPNSPP-Mish, and CD53s-CPANSPP-Leaky have the same batch 8 throughput with D53-FPNSPP-Leaky, but they respectively have 1\% and 1.6\% AP improvement with lower computing resources.  From the above improvement figures, we can see the huge advantages brought by model CSP-ization.  Therefore, we decided to use CD53s-CPANSPP-Mish, which results in the highest AP in Table \ref{table:t9} as the backbone of YOLOv4-CSP.

\begin{table}[t]
	\centering
	\begin{threeparttable}[h]
		\footnotesize
		\caption{Ablation study of partial at different position in computational block.}
		\label{table:t10}
		\setlength\tabcolsep{3.5pt}
		\begin{tabular}{lcccc}
			\toprule
			\textbf{Backbone} & \textbf{Neck} & \textbf{FLOPs} & \textbf{FPS$_{TX2}$} & \textbf{AP$^{val}$} \\				
			\midrule
			\textbf{tinyCD53s} & tinyFPN & 7.0B & 30 & 22.2\% \\
			\textbf{COSA-1x3x} & tinyFPN & 7.6B & 38 & \textbf{22.5\%} \\
			\textbf{COSA-2x2x} & tinyFPN & 6.9B & 42 & 22.0\% \\
			\textbf{COSA-3x1x} & tinyFPN & \textbf{6.3B} & \textbf{46} & 21.2\% \\
			\bottomrule
		\end{tabular}
	\end{threeparttable}
\end{table}

\begin{table}[t]
	\centering
	\begin{threeparttable}[h]
		\footnotesize
		\caption{Ablation study of training schedule with/without fine-tuning.}
		\label{table:t11}
		\setlength\tabcolsep{3.5pt}
		\begin{tabular}{lccccc}
			\toprule
			\textbf{Model} & \textbf{scratch} & \textbf{finetune} & \textbf{AP$^{val}$} & \textbf{AP$^{val}_{50}$} & \textbf{AP$^{val}_{75}$} \\				
			\midrule
			\textbf{YOLOv4-P5} & 300 & - & 50.5\% & 68.9\% & 55.2\% \\
			\textbf{YOLOv4-P5} & 300 & 150 & 51.7\% & 70.3\% & 56.7\% \\
			\textbf{YOLOv4-P6} & 300 & - & 53.4\% & 71.5\% & 58.5\% \\
			\textbf{YOLOv4-P6} & 300 & 150 & 54.4\% & 72.7\% & 59.5\% \\
			\textbf{YOLOv4-P7} & 300 & - & 54.6\% & 72.4\% & 59.7\% \\
			\textbf{YOLOv4-P7} & 300 & 150 & 55.3\% & 73.3\% & 60.4\% \\
			\bottomrule
		\end{tabular}
	\end{threeparttable}
\end{table}

\begin{table*}[h]
\centering
\begin{threeparttable}[h]
	\footnotesize
	\caption{Comparison of state-of-the-art object detectors.}
	\label{table:t12}
	\begin{tabular}{llllcccccc}
		\toprule
		\textbf{Method} & \textbf{Backbone} & \textbf{Size} & \textbf{FPS} & \textbf{AP} & \textbf{AP$_{50}$} & \textbf{AP$_{75}$} & \textbf{AP$_S$} & \textbf{AP$_M$} & \textbf{AP$_L$} \\	
		\midrule
		EfficientDet-D0 \cite{tan2019efficientdet} & EfficientNet-B0 \cite{tan2019efficientnet} & 512 & 97* & \textbf{34.6\%} & \textbf{53.0\%} & \textbf{37.1\%} & \textbf{12.4\%} & \textbf{39.0\%} & \textbf{52.7\%} \\
		\textbf{YOLOv4-CSP} & \textbf{CD53s} & 512 & 97/93* & \textbf{46.2\%} & \textbf{64.8\%} & \textbf{50.2\%} & \textbf{24.6\%} & \textbf{50.4\%} & \textbf{61.9\%} \\
		EfficientDet-D1 \cite{tan2019efficientdet} & EfficientNet-B1 \cite{tan2019efficientnet} & 640 & 74* & 40.5\% & 59.1\% & 43.7\% & 18.3\% & 45.0\% & 57.5\% \\
		\textbf{YOLOv4-CSP} & \textbf{CD53s} & 640 & 73/70* & \textbf{47.5\%} & \textbf{66.2\%} & \textbf{51.7\%} & \textbf{28.2\%} & \textbf{51.2\%} & \textbf{59.8\%} \\
		YOLOv3-SPP \cite{redmon2018yolov3} & D53 \cite{redmon2018yolov3} & 608 & 73 & 36.2\% & 60.6\% & 38.2\% & 20.6\% & 37.4\% & 46.1\% \\
		\textbf{YOLOv3-SPP ours} & D53 \cite{redmon2018yolov3} & 608 & 73 & 42.9\% & 62.4\% & 46.6\% & 25.9\% & 45.7\% & 52.4\% \\
		PP-YOLO \cite{long2020pp} & R50-vd-DCN \cite{long2020pp} & 608 & 73 & 45.2\% & 65.2\% & 49.9\% & 26.3\% & 47.8\% & 57.2\% \\
		YOLOv4 \cite{bochkovskiy2020yolov4} & CD53 \cite{bochkovskiy2020yolov4} & 608 & 62 & 43.5\% & 65.7\% & 47.3\% & 26.7\% & 46.7\% & 53.3\% \\
		\textbf{YOLOv4 ours} & CD53 \cite{bochkovskiy2020yolov4} & 608 & 62 & 45.5\% & 64.1\% & 49.5\% & 27.0\% & 49.0\% & 56.7\% \\
		\midrule
		EfficientDet-D2 \cite{tan2019efficientdet} & EfficientNet-B2 \cite{tan2019efficientnet} & 768 & 57* & 43.0\% & 62.3\% & 46.2\% & 22.5\% & 47.0\% & 58.4\% \\
		RetinaNet \cite{lin2017focal} & S49s \cite{du2019spinenet} & 640 & 53 & 41.5\% & 60.5\% & 44.6\% & 23.3\% & 45.0\% & 58.0\% \\
		ASFF \cite{liu2019learning} & D53 \cite{redmon2018yolov3} & 608* & 46 & 42.4\% & 63.0\% & 47.4\% & 25.5\% & 45.7\% & 52.3\% \\
		\textbf{YOLOv4-P5} & \textbf{CSP-P5} & 896 & 43/41* & \textbf{51.8\%} & \textbf{70.3\%} & \textbf{56.6\%} & \textbf{33.4\%} & \textbf{55.7\%} & \textbf{63.4\%} \\
		RetinaNet \cite{lin2017focal} & S49 \cite{du2019spinenet} & 640 & 42 & 44.3\% & 63.8\% & 47.6\% & 25.9\% & 47.7\% & 61.1\% \\
		EfficientDet-D3 \cite{tan2019efficientdet} & EfficientNet-B3 \cite{tan2019efficientnet} & 896 & 36* & 47.5\% & 66.2\% & 51.5\% & 27.9\% & 51.4\% & 62.0\% \\
		\textbf{YOLOv4-P6} & \textbf{CSP-P6} & 1280 & 32/30* & \textbf{54.5\%} & \textbf{72.6\%} & \textbf{59.8\%} & \textbf{36.8\%} & \textbf{58.3\%} & \textbf{65.9\%} \\
		\midrule
		ASFF\cite{liu2019learning} & D53 \cite{redmon2018yolov3} & 800* & 29 & 43.9\% & 64.1\% & 49.2\% & 27.0\% & 46.6\% & 53.4\% \\
		SM-NAS: E2 \cite{yao2019sm} & - & 800*600 & 25 & 40.0\% & 58.2\% & 43.4\% & 21.1\% & 42.4\% & 51.7\% \\
		EfficientDet-D4 \cite{tan2019efficientdet} & EfficientNet-B4 \cite{tan2019efficientnet} & 1024 & 23* & 49.7\% & 68.4\% & 53.9\% & 30.7\% & 53.2\% & 63.2\% \\
		SM-NAS: E3 \cite{yao2019sm} & - & 800*600 & 20 & 42.8\% & 61.2\% & 46.5\% & 23.5\% & 45.5\% & 55.6\% \\
		RetinaNet \cite{lin2017focal} & S96 \cite{du2019spinenet} & 1024 & 19 & 48.6\% & 68.4\% & 52.5\% & 32.0\% & 52.3\% & 62.0\% \\
		ATSS \cite{zhang2019bridging} & R101 \cite{he2016deep} & 800* & 18 & 43.6\% & 62.1\% & 47.4\% & 26.1\% & 47.0\% & 53.6\% \\
		\textbf{YOLOv4-P7} & \textbf{CSP-P7} & 1536 & 17/16* & \textbf{55.5\%} & \textbf{73.4\%} & \textbf{60.8\%} & \textbf{38.4\%} & \textbf{59.4\%} & \textbf{67.7\%} \\
		RDSNet \cite{wang2019rdsnet} & R101 \cite{he2016deep} & 600 & 17 & 36.0\% & 55.2\% & 38.7\% & 17.4\% & 39.6\% & 49.7\% \\
		CenterMask \cite{lee2019centermask} & R101-FPN \cite{lin2017feature} & - & 15 & 44.0\% & - & - & 25.8\% & 46.8\% & 54.9\% \\
		\midrule
		EfficientDet-D5 \cite{tan2019efficientdet} & EfficientNet-B5 \cite{tan2019efficientnet} & 1280 & 14* & 51.5\% & 70.5\% & 56.7\% & 33.9\% & 54.7\% & 64.1\% \\
		ATSS \cite{zhang2019bridging} & R101-DCN \cite{dai2017deformable} & 800* & 14 & 46.3\% & 64.7\% & 50.4\% & 27.7\% & 49.8\% & 58.4\% \\
		SABL \cite{wang2020side} & R101 \cite{he2016deep} & - & 13 & 43.2\% & 62.0\% & 46.6\% & 25.7\% & 47.4\% & 53.9\% \\
		CenterMask \cite{lee2019centermask} & V99-FPN \cite{lee2019centermask} & - & 13 & 46.5\% & - & - & 28.7\% & 48.9\% & 57.2\% \\
		EfficientDet-D6 \cite{tan2019efficientdet} & EfficientNet-B6 \cite{tan2019efficientnet} & 1408 & 11* & 52.6\% & 71.5\% & 57.2\% & 34.9\% & 56.0\% & 65.4\% \\
		RDSNet \cite{wang2019rdsnet} & R101 \cite{he2016deep} & 800 & 11 & 38.1\% & 58.5\% & 40.8\% & 21.2\% & 41.5\% & 48.2\% \\
		RetinaNet \cite{lin2017focal} & S143 \cite{du2019spinenet} & 1280 & 10 & 50.7\% & 70.4\% & 54.9\% & 33.6\% & 53.9\% & 62.1\% \\
		SM-NAS: E5 \cite{yao2019sm} & - & 1333*800 & 9.3 & 45.9\% & 64.6\% & 49.6\% & 27.1\% & 49.0\% & 58.0\% \\
		EfficientDet-D7 \cite{tan2019efficientdet} & EfficientNet-B6 \cite{tan2019efficientnet} & 1536 & 8.2* & 53.7\% & 72.4\% & 58.4\% & 35.8\% & 57.0\% & 66.3\% \\
		ATSS \cite{zhang2019bridging} & X-32x8d-101-DCN \cite{dai2017deformable} & 800* & 7.0 & 47.7\% & 66.6\% & 52.1\% & 29.3\% & 50.8\% & 59.7\% \\
		ATSS \cite{zhang2019bridging} & X-64x4d-101-DCN \cite{dai2017deformable} & 800* & 6.9 & 47.7\% & 66.5\% & 51.9\% & 29.7\% & 50.8\% & 59.4\% \\
		EfficientDet-D7x \cite{tan2019efficientdet} & EfficientNet-B7 \cite{tan2019efficientnet} & 1536 & 6.5* & 55.1\% & \textbf{74.3\%} & 59.9\% & 37.2\% & 57.9\% & \textbf{68.0\%} \\
		TSD \cite{song2020revisiting} & R101 \cite{he2016deep} & - & 5.3* & 43.2\% & 64.0\% & 46.9\% & 24.0\% & 46.3\% & 55.8\% \\
		\bottomrule
	\end{tabular}
\end{threeparttable}
\end{table*}

\subsection{Ablation study on YOLOv4-tiny}

In this sub-section, we design an experiment to show how flexible can be if one uses CSPNet with partial functions in computational blocks.  We also compare with CSPDarknet53, in which we perform linear scaling down on width and depth.  The results are shown in Table \ref{table:t10}.

From the figures shown in Table \ref{table:t10}, we can see that the designed PCB technique can make the model more flexible, because such a design can be adjusted according to actual needs.  From the above results, we also confirmed that linear scaling down does have its limitation.  It is apparent that when under limited operating conditions, the residual addition of tinyCD53s becomes the bottleneck of inference speed, because its frame rate is much lower than the COSA architecture with the same amount of computations.  Meanwhile, we also see that the proposed COSA can get a higher AP.  Therefore, we finally chose COSA-2x2x which received the best speed/accuracy trade-off in our experiment as the YOLOv4-tiny architecture.

\subsection{Ablation study on YOLOv4-large}

In Table \ref{table:t11} we show the AP obtained by YOLOv4 models in training from scratch and fine-tune stages. 

\subsection{Scaled-YOLOv4 for object detection}

We compare with other real-time object detectors, and the results are shown in Table \ref{table:t12}.  The values marked in bold in the [AP, AP$_{50}$, AP$_{75}$, AP$_S$, AP$_M$, AP$_L$] items indicate that model is the best performer in the corresponding item.  We can see that all scaled YOLOv4 models, including YOLOv4-CSP, YOLOv4-P5, YOLOv4-P6, YOLOv4-P7, are Pareto optimal on all indicators.  When we compare YOLOv4-CSP with the same accuracy of EfficientDet-D3 (47.5\% vs 47.5\%), the inference speed is 1.9 times.  When YOLOv4-P5 is compared with EfficientDet-D5 with the same accuracy (51.8\% vs 51.5\%), the inference speed is 2.9 times.  The situation is similar to the comparisons between YOLOv4-P6 vs EfficientDet-D7 (54.5\% vs 53.7\%) and YOLOv4-P7 vs EfficientDet-D7x (55.5\% vs 55.1\%).  In both cases, YOLOv4-P6 and YOLOv4-P7 are, respectively, 3.7 times and 2.5 times faster in terms of inference speed.  All scaled-YOLOv4 models reached state-of-the-art results.

The results of test-time augmentation (TTA) experiments of YOLOv4-large models are shown in Table \ref{table:t13}. YOLOv4-P5, YOLOv4-P6, and YOLOv4-P7 gets 1.1\%, 0.7\%, and 0.5\% higher AP, respectively, after TTA is applied.

\begin{table}[h]
	\centering
	\begin{threeparttable}[h]
		\footnotesize
		\caption{Results of YOLOv4-large models with test-time augmentation (TTA).}
		\label{table:t13}
		\setlength\tabcolsep{3.5pt}
		\begin{tabular}{lccc}
			\toprule
			\textbf{Model} & \textbf{AP} & \textbf{AP$_{50}$} & \textbf{AP$_{75}$} \\				
			\midrule
			\textbf{YOLOv4-P5} with TTA & 52.9\% & 70.7\% & 58.3\% \\
			\textbf{YOLOv4-P6} with TTA & 55.2\% & 72.9\% & 60.5\% \\
			\textbf{YOLOv4-P7} with TTA & 56.0\% & 73.3\% & 61.4\% \\
			\bottomrule
		\end{tabular}
	\end{threeparttable}
\end{table}

We then compare the performance of YOLOv4-tiny with that of other tiny object detectors, and the results are shown in Table \ref{table:t14}.  It is apparent that YOLOv4-tiny achieves the best performance in comparison with other tiny models.

\begin{table}[h]
	\centering
	\begin{threeparttable}[h]
		\footnotesize
		\caption{Comparison of state-of-the-art tiny models.}
		\label{table:t14}
		\setlength\tabcolsep{3.5pt}
		\begin{tabular}{lcccc}
			\toprule
			\textbf{Model} & \textbf{Size} & \textbf{FPS$_{1080ti}$} & \textbf{FPS$_{TX2}$} & \textbf{AP} \\				
			\midrule
			\textbf{YOLOv4-tiny} & 416 & \textbf{371} & \textbf{42} & 21.7\% \\
			\textbf{YOLOv4-tiny (3l)} & 320 & 252 & 41 & \textbf{28.7\%} \\
			ThunderS146 \cite{qin2019thundernet} & 320 & 248 & - & 23.6\% \\
			%ThunderS535 \cite{qin2019thundernet} & 320 & 214 & - & 28.0\% \\
			CSPPeleeRef \cite{wang2020cspnet} & 320 & 205 & 41 & 23.5\% \\
			YOLOv3-tiny \cite{redmon2018yolov3} & 416 & 368 & 37 & 16.6\% \\
			\bottomrule
		\end{tabular}
	\end{threeparttable}
\end{table}

Finally, we put YOLOv4-tiny on different embedded GPUs for testing, including Xavier AGX, Xavier NX, Jetson TX2, Jetson NANO.  We also use TensorRT FP32 (FP16 if supported) for testing. All frame rates obtained by different models are listed in Table \ref{table:t15}.  It is apparent that YOLOv4-tiny can achieve real-time performance no matter which device is used.  If we adopt FP16 and batch size 4 to test Xavier AGX and Xavier NX, the frame rate can reach 380 FPS and 199 FPS respectively.  In addition, if one uses TensorRT FP16 to run YOLOv4-tiny on general GPU RTX 2080ti, when the batch size respectively equals to 1 and 4, the respective frame rate can reach 773 FPS and 1774 FPS, which is extremely fast.

\begin{table}[h]
	\centering
	\begin{threeparttable}[h]
		\footnotesize
		\caption{FPS of YOLOv4-tiny on embedded devices.}
		\label{table:t15}
		\setlength\tabcolsep{3.5pt}
		\begin{tabular}{lcccc}
			\toprule
			\textbf{TensorRT.} & \textbf{FPS$_{AGX}$} & \textbf{FPS$_{NX}$} & \textbf{FPS$_{TX2}$} & \textbf{FPS$_{NANO}$} \\				
			\midrule
			without & 120 & 75 & 42 & 16 \\
			with & 290 & 118 & 100 & 39 \\
			\bottomrule
		\end{tabular}
	\end{threeparttable}
\end{table}

\subsection{Scaled-YOLOv4 as na\"ive once-for-all model}

\begin{figure}[h]
	\begin{center}
		\includegraphics[width=1.0\linewidth]{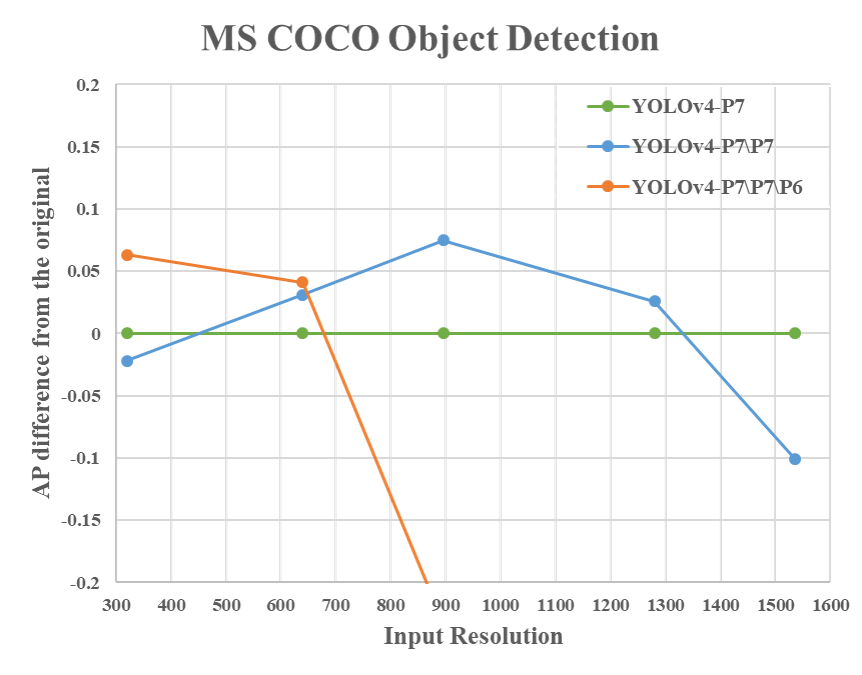}
	\end{center}
	\vspace{-4mm}
	\caption{YOLOv4-P7 as ``once-for-all'' model.}
	\vspace{4mm}
	\label{fig:ofa}
\end{figure}

In this sub-section, we design experiments to show that an FPN-like architecture is a naïve once-for-all model. Here we remove some stages of top-down path and detection branch of YOLOv4-P7. YOLOv4-P7$\backslash$P7 and YOLOv4-P7$\backslash$P7$\backslash$P6 represent the model which has removed \{P7\} and \{P7, P6\} stages from the trained YOLOv4-P7. Figure \ref{fig:ofa} shows the AP difference between pruned models and original YOLOv4-P7 with different input resolution.

We can find that YOLOv4-P7 has the best AP at high resolution, while YOLOv4-P7$\backslash$P7 and YOLOv4-P7$\backslash$P7$\backslash$P6 have the best AP at middle and low resolution, respectively. This means that we can use sub-nets of FPN-like models to execute the object detection task well. Moreover, we can perform compound scale-down the model architectures and input size of an object detector to get the best performance.

%-------------------------------------------------------------------------
\section{Conclusions}

We show that the YOLOv4 object detection neural network based on the CSP approach, scales both up and down and is applicable to small and large networks. So we achieve the highest accuracy 56.0\% AP on test-dev COCO dataset for the model YOLOv4-large, extremely high speed 1774 FPS for the small model YOLOv4-tiny on RTX 2080Ti by using TensorRT-FP16, and optimal speed and accuracy for other YOLOv4 models.

\vspace{6mm}

%------------------------------------------------------------------------
\section{Acknowledgements}

The authors wish to thank National Center for High-performance Computing (NCHC) for
providing computational and storage resources. A large part of the code is borrowed from \url{https://github.com/AlexeyAB}, \url{https://github.com/WongKinYiu} and \url{https://github.com/glenn-jocher}. Thanks for their wonderful works.

%------------------------------------------------------------------------

%\end{CJK*}

\clearpage
\clearpage
\clearpage
{\small
\begin{thebibliography}{10}\itemsep=-1pt
	
	\bibitem{bochkovskiy2020yolov4}
	Alexey Bochkovskiy, Chien-Yao Wang, and Hong-Yuan~Mark Liao.
	\newblock {YOLOv4}: Optimal speed and accuracy of object detection.
	\newblock {\em arXiv preprint arXiv:2004.10934}, 2020.
	
	\bibitem{cai2019once}
	Han Cai, Chuang Gan, Tianzhe Wang, Zhekai Zhang, and Song Han.
	\newblock Once-for-all: Train one network and specialize it for efficient
	deployment.
	\newblock {\em arXiv preprint arXiv:1908.09791}, 2019.
	
	\bibitem{cao2020d2det}
	Jiale Cao, Hisham Cholakkal, Rao~Muhammad Anwer, Fahad~Shahbaz Khan, Yanwei
	Pang, and Ling Shao.
	\newblock {D2Det}: Towards high quality object detection and instance
	segmentation.
	\newblock In {\em Proceedings of the IEEE Conference on Computer Vision and
		Pattern Recognition (CVPR)}, pages 11485--11494, 2020.
	
	\bibitem{chao2019hardnet}
	Ping Chao, Chao-Yang Kao, Yu-Shan Ruan, Chien-Hsiang Huang, and Youn-Long Lin.
	\newblock {HarDNet}: A low memory traffic network.
	\newblock {\em Proceedings of the IEEE International Conference on Computer
		Vision (ICCV)}, 2019.
	
	\bibitem{dai2017deformable}
	Jifeng Dai, Haozhi Qi, Yuwen Xiong, Yi Li, Guodong Zhang, Han Hu, and Yichen
	Wei.
	\newblock Deformable convolutional networks.
	\newblock In {\em Proceedings of the IEEE International Conference on Computer
		Vision (ICCV)}, pages 764--773, 2017.
	
	\bibitem{du2019spinenet}
	Xianzhi Du, Tsung-Yi Lin, Pengchong Jin, Golnaz Ghiasi, Mingxing Tan, Yin Cui,
	Quoc~V Le, and Xiaodan Song.
	\newblock {SpineNet}: Learning scale-permuted backbone for recognition and
	localization.
	\newblock {\em arXiv preprint arXiv:1912.05027}, 2019.
	
	\bibitem{duan2019centernet}
	Kaiwen Duan, Song Bai, Lingxi Xie, Honggang Qi, Qingming Huang, and Qi Tian.
	\newblock {CenterNet}: Keypoint triplets for object detection.
	\newblock In {\em Proceedings of the IEEE International Conference on Computer
		Vision (ICCV)}, pages 6569--6578, 2019.
	
	\bibitem{ghiasi2019fpn}
	Golnaz Ghiasi, Tsung-Yi Lin, and Quoc~V Le.
	\newblock {NAS-FPN}: Learning scalable feature pyramid architecture for object
	detection.
	\newblock In {\em Proceedings of the IEEE Conference on Computer Vision and
		Pattern Recognition (CVPR)}, pages 7036--7045, 2019.
	
	\bibitem{girshick2015fast}
	Ross Girshick.
	\newblock {Fast R-CNN}.
	\newblock In {\em Proceedings of the IEEE International Conference on Computer
		Vision (ICCV)}, pages 1440--1448, 2015.
	
	\bibitem{girshick2014rich}
	Ross Girshick, Jeff Donahue, Trevor Darrell, and Jitendra Malik.
	\newblock Rich feature hierarchies for accurate object detection and semantic
	segmentation.
	\newblock In {\em Proceedings of the IEEE Conference on Computer Vision and
		Pattern Recognition (CVPR)}, pages 580--587, 2014.
	
	\bibitem{he2016deep}
	Kaiming He, Xiangyu Zhang, Shaoqing Ren, and Jian Sun.
	\newblock Deep residual learning for image recognition.
	\newblock In {\em Proceedings of the IEEE Conference on Computer Vision and
		Pattern Recognition (CVPR)}, pages 770--778, 2016.
	
	\bibitem{huang2017densely}
	Gao Huang, Zhuang Liu, Laurens Van Der~Maaten, and Kilian~Q Weinberger.
	\newblock Densely connected convolutional networks.
	\newblock In {\em Proceedings of the IEEE Conference on Computer Vision and
		Pattern Recognition (CVPR)}, pages 4700--4708, 2017.
	
	\bibitem{law2018cornernet}
	Hei Law and Jia Deng.
	\newblock {CornerNet}: Detecting objects as paired keypoints.
	\newblock In {\em Proceedings of the European Conference on Computer Vision
		(ECCV)}, pages 734--750, 2018.
	
	\bibitem{law2019cornernet}
	Hei Law, Yun Teng, Olga Russakovsky, and Jia Deng.
	\newblock {CornerNet-Lite}: Efficient keypoint based object detection.
	\newblock {\em arXiv preprint arXiv:1904.08900}, 2019.
	
	\bibitem{lee2019energy}
	Youngwan Lee, Joong-won Hwang, Sangrok Lee, Yuseok Bae, and Jongyoul Park.
	\newblock An energy and {GPU-computation} efficient backbone network for
	real-time object detection.
	\newblock In {\em Proceedings of the IEEE Conference on Computer Vision and
		Pattern Recognition Workshop (CVPR Workshop)}, 2019.
	
	\bibitem{lee2019centermask}
	Youngwan Lee and Jongyoul Park.
	\newblock {CenterMask}: Real-time anchor-free instance segmentation.
	\newblock In {\em Proceedings of the IEEE Conference on Computer Vision and
		Pattern Recognition (CVPR)}, 2020.
	
	\bibitem{lin2017feature}
	Tsung-Yi Lin, Piotr Doll{\'a}r, Ross Girshick, Kaiming He, Bharath Hariharan,
	and Serge Belongie.
	\newblock Feature pyramid networks for object detection.
	\newblock In {\em Proceedings of the IEEE Conference on Computer Vision and
		Pattern Recognition (CVPR)}, pages 2117--2125, 2017.
	
	\bibitem{lin2017focal}
	Tsung-Yi Lin, Priya Goyal, Ross Girshick, Kaiming He, and Piotr Doll{\'a}r.
	\newblock Focal loss for dense object detection.
	\newblock In {\em Proceedings of the IEEE International Conference on Computer
		Vision (ICCV)}, pages 2980--2988, 2017.
	
	\bibitem{liu2019learning}
	Songtao Liu, Di Huang, and Yunhong Wang.
	\newblock Learning spatial fusion for single-shot object detection.
	\newblock {\em arXiv preprint arXiv:1911.09516}, 2019.
	
	\bibitem{liu2018path}
	Shu Liu, Lu Qi, Haifang Qin, Jianping Shi, and Jiaya Jia.
	\newblock Path aggregation network for instance segmentation.
	\newblock In {\em Proceedings of the IEEE Conference on Computer Vision and
		Pattern Recognition (CVPR)}, pages 8759--8768, 2018.
	
	\bibitem{liu2016ssd}
	Wei Liu, Dragomir Anguelov, Dumitru Erhan, Christian Szegedy, Scott Reed,
	Cheng-Yang Fu, and Alexander~C Berg.
	\newblock {SSD}: Single shot multibox detector.
	\newblock In {\em Proceedings of the European Conference on Computer Vision
		(ECCV)}, pages 21--37, 2016.
	
	\bibitem{long2020pp}
	Xiang Long, Kaipeng Deng, Guanzhong Wang, Yang Zhang, Qingqing Dang, Yuan Gao,
	Hui Shen, Jianguo Ren, Shumin Han, Errui Ding, et~al.
	\newblock {PP-YOLO}: An effective and efficient implementation of object
	detector.
	\newblock {\em arXiv preprint arXiv:2007.12099}, 2020.
	
	\bibitem{ma2018shufflenetv2}
	Ningning Ma, Xiangyu Zhang, Hai-Tao Zheng, and Jian Sun.
	\newblock {ShuffleNetV2}: Practical guidelines for efficient cnn architecture
	design.
	\newblock In {\em Proceedings of the European Conference on Computer Vision
		(ECCV)}, pages 116--131, 2018.
	
	\bibitem{qiao2020detectors}
	Siyuan Qiao, Liang-Chieh Chen, and Alan Yuille.
	\newblock {DetectoRS}: Detecting objects with recursive feature pyramid and
	switchable atrous convolution.
	\newblock {\em arXiv preprint arXiv:2006.02334}, 2020.
	
	\bibitem{qin2019thundernet}
	Zheng Qin, Zeming Li, Zhaoning Zhang, Yiping Bao, Gang Yu, Yuxing Peng, and
	Jian Sun.
	\newblock {ThunderNet}: Towards real-time generic object detection.
	\newblock {\em Proceedings of the IEEE International Conference on Computer
		Vision (ICCV)}, 2019.
	
	\bibitem{qiu2020borderdet}
	Han Qiu, Yuchen Ma, Zeming Li, Songtao Liu, and Jian Sun.
	\newblock {BorderDet}: Border feature for dense object detection.
	\newblock In {\em Proceedings of the European Conference on Computer Vision
		(ECCV)}, pages 549--564. Springer, 2020.
	
	\bibitem{radosavovic2020designing}
	Ilija Radosavovic, Raj~Prateek Kosaraju, Ross Girshick, Kaiming He, and Piotr
	Doll{\'a}r.
	\newblock Designing network design spaces.
	\newblock In {\em Proceedings of the IEEE Conference on Computer Vision and
		Pattern Recognition (CVPR)}, pages 10428--10436, 2020.
	
	\bibitem{redmon2016you}
	Joseph Redmon, Santosh Divvala, Ross Girshick, and Ali Farhadi.
	\newblock You only look once: Unified, real-time object detection.
	\newblock In {\em Proceedings of the IEEE Conference on Computer Vision and
		Pattern Recognition (CVPR)}, pages 779--788, 2016.
	
	\bibitem{redmon2017yolo9000}
	Joseph Redmon and Ali Farhadi.
	\newblock {YOLO9000}: better, faster, stronger.
	\newblock In {\em Proceedings of the IEEE Conference on Computer Vision and
		Pattern Recognition (CVPR)}, pages 7263--7271, 2017.
	
	\bibitem{redmon2018yolov3}
	Joseph Redmon and Ali Farhadi.
	\newblock {YOLOv3}: An incremental improvement.
	\newblock {\em arXiv preprint arXiv:1804.02767}, 2018.
	
	\bibitem{ren2015faster}
	Shaoqing Ren, Kaiming He, Ross Girshick, and Jian Sun.
	\newblock {Faster R-CNN}: Towards real-time object detection with region
	proposal networks.
	\newblock In {\em Advances in Neural Information Processing Systems (NIPS)},
	pages 91--99, 2015.
	
	\bibitem{simonyan2014very}
	Karen Simonyan and Andrew Zisserman.
	\newblock Very deep convolutional networks for large-scale image recognition.
	\newblock {\em arXiv preprint arXiv:1409.1556}, 2014.
	
	\bibitem{song2020revisiting}
	Guanglu Song, Yu Liu, and Xiaogang Wang.
	\newblock Revisiting the sibling head in object detector.
	\newblock In {\em Proceedings of the IEEE Conference on Computer Vision and
		Pattern Recognition (CVPR)}, pages 11563--11572, 2020.
	
	\bibitem{tan2019efficientnet}
	Mingxing Tan and Quoc~V Le.
	\newblock {EfficientNet}: Rethinking model scaling for convolutional neural
	networks.
	\newblock In {\em Proceedings of International Conference on Machine Learning
		(ICML)}, 2019.
	
	\bibitem{tan2019efficientdet}
	Mingxing Tan, Ruoming Pang, and Quoc~V Le.
	\newblock {EfficientDet}: Scalable and efficient object detection.
	\newblock In {\em Proceedings of the IEEE Conference on Computer Vision and
		Pattern Recognition (CVPR)}, 2020.
	
	\bibitem{tian2019fcos}
	Zhi Tian, Chunhua Shen, Hao Chen, and Tong He.
	\newblock {FCOS}: Fully convolutional one-stage object detection.
	\newblock In {\em Proceedings of the IEEE International Conference on Computer
		Vision (ICCV)}, pages 9627--9636, 2019.
	
	\bibitem{wang2020cspnet}
	Chien-Yao Wang, Hong-Yuan~Mark Liao, Yueh-Hua Wu, Ping-Yang Chen, Jun-Wei
	Hsieh, and I-Hau Yeh.
	\newblock {CSPNet}: A new backbone that can enhance learning capability of
	{CNN}.
	\newblock {\em Proceedings of the IEEE Conference on Computer Vision and
		Pattern Recognition Workshop (CVPR Workshop)}, 2020.
	
	\bibitem{wang2020side}
	Jiaqi Wang, Wenwei Zhang, Yuhang Cao, Kai Chen, Jiangmiao Pang, Tao Gong,
	Jianping Shi, Chen~Change Loy, and Dahua Lin.
	\newblock Side-aware boundary localization for more precise object detection.
	\newblock In {\em Proceedings of the European Conference on Computer Vision
		(ECCV)}, pages 403--419. Springer, 2020.
	
	\bibitem{wang2019rdsnet}
	Shaoru Wang, Yongchao Gong, Junliang Xing, Lichao Huang, Chang Huang, and
	Weiming Hu.
	\newblock {RDSNet}: A new deep architecture for reciprocal object detection and
	instance segmentation.
	\newblock {\em arXiv preprint arXiv:1912.05070}, 2019.
	
	\bibitem{wang2020scale}
	Xinjiang Wang, Shilong Zhang, Zhuoran Yu, Litong Feng, and Wayne Zhang.
	\newblock Scale-equalizing pyramid convolution for object detection.
	\newblock In {\em Proceedings of the IEEE Conference on Computer Vision and
		Pattern Recognition (CVPR)}, pages 13359--13368, 2020.
	
	\bibitem{xie2017aggregated}
	Saining Xie, Ross Girshick, Piotr Doll{\'a}r, Zhuowen Tu, and Kaiming He.
	\newblock Aggregated residual transformations for deep neural networks.
	\newblock In {\em Proceedings of the IEEE Conference on Computer Vision and
		Pattern Recognition (CVPR)}, pages 1492--1500, 2017.
	
	\bibitem{yao2019sm}
	Lewei Yao, Hang Xu, Wei Zhang, Xiaodan Liang, and Zhenguo Li.
	\newblock {SM-NAS}: Structural-to-modular neural architecture search for object
	detection.
	\newblock In {\em Proceedings of the AAAI Conference on Artificial Intelligence
		(AAAI)}, 2020.
	
	\bibitem{zagoruyko2016wide}
	Sergey Zagoruyko and Nikos Komodakis.
	\newblock Wide residual networks.
	\newblock {\em arXiv preprint arXiv:1605.07146}, 2016.
	
	\bibitem{zhang2020dynamic}
	Hongkai Zhang, Hong Chang, Bingpeng Ma, Naiyan Wang, and Xilin Chen.
	\newblock {Dynamic R-CNN}: Towards high quality object detection via dynamic
	training.
	\newblock In {\em Proceedings of the European Conference on Computer Vision
		(ECCV)}, pages 260--275. Springer, 2020.
	
	\bibitem{zhang2019bridging}
	Shifeng Zhang, Cheng Chi, Yongqiang Yao, Zhen Lei, and Stan~Z Li.
	\newblock Bridging the gap between anchor-based and anchor-free detection via
	adaptive training sample selection.
	\newblock In {\em Proceedings of the IEEE Conference on Computer Vision and
		Pattern Recognition (CVPR)}, 2020.
	
	\bibitem{zhou2019objects}
	Xingyi Zhou, Dequan Wang, and Philipp Kr{\"a}henb{\"u}hl.
	\newblock Objects as points.
	\newblock In {\em arXiv preprint arXiv:1904.07850}, 2019.
	
\end{thebibliography}

}
\end{document}